\documentclass{article}

\usepackage{graphicx}
\usepackage{amsmath}
\usepackage[preprint]{neurips_2025}

\usepackage{listings}

\usepackage[utf8]{inputenc} 
\usepackage[T1]{fontenc}	
\usepackage{hyperref}   	
\usepackage{url}        	
\usepackage{booktabs}   	
\usepackage{amsfonts}   	
\usepackage{nicefrac}   	
\usepackage{microtype}  	
\usepackage{xcolor}     	

\title{Secu-Table: a Comprehensive security table dataset for evaluating semantic table interpretation systems}

\author{
    Azanzi Jiomekong\\
    Department of Computer Science, University of Yaounde I\\TIB Leibniz Information Centre for Science and Technology
    \And
    Jean Bikim\\
    Department of Computer Science, University of Yaounde I\\TIB Leibniz Information Centre for Science and Technology
    \And
    Patricia Negoue\\
    Department of Computer Science, University of Yaounde I
    \And
    Joyce Chin\\
    Department of Computer Science, University of Yaounde I
}

\begin{document}

\maketitle

\begin{abstract}

Evaluating semantic tables interpretation (STI) systems, (particularly, those based on Large Language Models- LLMs) especially in domain-specific contexts such as the security domain, depends heavily on the dataset. However, in the security domain, tabular datasets for state-of-the-art are not publicly available. In this paper, we introduce Secu-Table dataset, composed of more than 1500 tables with more than 15k entities constructed using security data extracted from Common Vulnerabilities and Exposures (CVE) and Common Weakness Enumeration (CWE) data sources and annotated using Wikidata and the SEmantic Processing of Security Event Streams CyberSecurity Knowledge Graph (SEPSES CSKG). Along with the dataset, all the code is publicly released. This dataset is made available to the research community in the context of the SemTab challenge on Tabular to Knowledge Graph Matching. This challenge aims to evaluate the performance of several STI based on open source LLMs. Preliminary evaluation, serving as baseline, was conducted using Falcon3-7b-instruct and Mistral-7B-Instruct, two open source LLMs and GPT-4o mini one closed source LLM.
\end{abstract}

\section{Background \& Summary}
\label{backgroundSummary}
In the field of cybersecurity, several datasets such as Common Vulnerabilities and Exposures (CVE)~\cite{vulnerabilities2005common}, Common Attack Pattern Enumeration and Classification (CAPEC)~\cite{barnum2008common}, Common Weakness Enumeration (CWE)~\cite{christey2013common}, etc. has been released for various purposes such as development, testing, education, etc. These datasets are used for enabling academics and security professionals to study attack patterns, vulnerabilities, and defence mechanisms; provides data for training and evaluating machine learning (ML) models used in intrusion detection systems, malware analysis, and threat intelligence platforms; offer realistic or synthetic data to test the effectiveness of security tools and techniques~\cite{abrar2020machine,belgrana2021network,oyama2019identifying,csandor2023ember,vasan2020imcfn,choudhary2023machine,dutta2021detecting,castano2023phikita,yu2023honey,dani2024ai,blocki2023towards}.

Security datasets often suffer from limitations. On one hand, Security datasets are scattered on the Internet and provided in heterogeneous formats such as CSV, JSON, XSL, or XML formats. This makes it difficult to get a holistic view of the interconnectedness of information across different data sources. On the other hand, many datasets focus on specific attack vectors or limited environments, limiting generalisability; There is a lack of detailed annotations in datasets, making it difficult to train supervised learning models. To solve these limits, security data can be extracted from diverse data sources, organised using a tabular data format and linked to existing knowledge graphs (KGs). This is called Semantic Table Interpretation~\cite{efthymiou2017matching,Jiomekong2022TSOTSATable}. The KGs schema will help align different terminologies and understand the relationships between concepts.

Although humans can manually annotate tabular data, understanding the semantics of tables and annotating large volumes of data remains complex, resource-heavy and time-consuming~\cite{jiomekong2019extracting}. This has led to scientific challenges such as Tabular Data to Knowledge Graph Challenge Matching or SemTab challenge~\cite{efthymiou2017matching,Foko2023NaiveBayes,Bikim2024SemTab}\footnote{\url{https://www.cs.ox.ac.uk/isg/challenges/sem-tab/}}. Launched in 2019 and hosted by the International Semantic Web Challenge (ISWC), this challenge aims to benchmark systems dealing with the problem of matching tabular data to KGs. This consists of linking table elements (such as entities in cells, column types, relations between columns) to their corresponding entities in the KG. Although several tabular datasets have been proposed~\cite{Hulsebos2021GitTablesAL,BiodivTab2021,Jiomekong2022FCT,ToughTablesCutrona2020}, datasets specific in the domain of security are not yet publicly available.

This paper contributes to the SemTab community and the SemTab@ISWC challenge\footnote{\url{https://sem-tab-challenge.github.io/2025/}} by introducing the secu-table dataset, composed of tables extracted from CVE and CWE data sources and annotated using Wikidata~\cite{wikidataVrandevcic2014} and SEmantic Processing of Security Event Streams CyberSecurity Knowledge Graph (SEPSES CSKG)~\cite{inbook_SEPSESKGs}. Given that the dataset integrates data extracted from open cybersecurity resources, and knowledge graphs published under the Creative Commons Zero (CC0 1.0 Universal Public Domain Dedication) and the Creative Commons Attribution 4.0 International (CC BY 4.0) licenses, the secu-table dataset released in this work adopted the CC BY 4.0 license to respect the source licenses. The dataset is available at \url{https://huggingface.co/datasets/jiofidelus/SecuTable}.

The current version of the secu-table dataset (available at \url{https://huggingface.co/datasets/jiofidelus/SecuTable/tree/main/secutable\_v2}) is composed of more than 1500 tables. These tables contain more than 150k entities, more than 1M lines and more than 20k columns. The average number of columns per table is 8.13 and the average number of rows per table is 291.63. It was made available to the research community in the context of SemTab@ISWC 2025 challenge on Tabular to Knowledge Graph Matching hosted by the 24th International Semantic Web Conference\footnote{\url{https://sem-tab-challenge.github.io/2025/}}. It aims at evaluating STI systems, particularly those based on large language models (LLMs). 

From January onward, new releases of the dataset will occur on a quarterly basis, with expanded coverage from security data sources such as Adversarial Tactics, Techniques, and Common Knowledge (ATT\&CK); Common Configuration Enumeration (CCE); Common Platform Enumeration (CPE); Common Vulnerability Scoring System (CSVS); Open Worldwide Application Security Project (OWASP); Security Content Automation Protocol (SCAP); etc.

\section{Methods}
\label{sec:methodology}
\begin{figure}[!ht]
    \centering
    \includegraphics[scale=0.3]{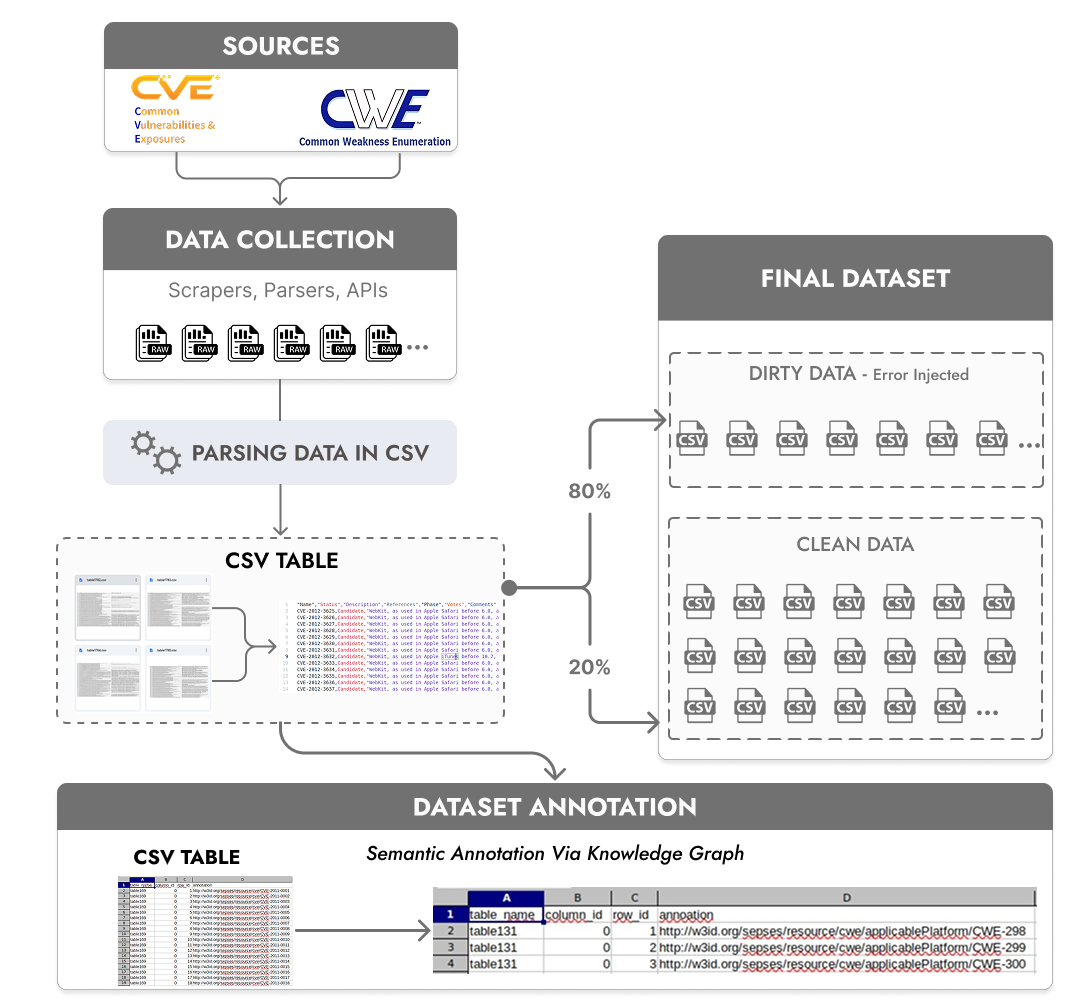}
    \caption{Secu-Table construction pipeline}
    \label{fig:secuTablePipeline}
\end{figure}

This section presents the input data used to create the secu-table dataset and step by step construction method. The first two steps of the method consisted of the recruitment of the data curators and the identification of data sources. Thereafter, the construction pipeline presented by Fig. \ref{fig:secuTablePipeline} was executed.

\subsection{Input Data}
The input data used to construct the current version of the secu-table dataset were derived from the Common Vulnerability and Exposures\footnote{\url{https://www.cve.org/}} (CVE)~\cite{vulnerabilities2005common} and Common Weakness Enumeration\footnote{\url{https://cwe.mitre.org/}} (CWE)~\cite{christey2013common} downloaded in 2022. The dataset obtained was annotated using Wikidata~\footnote{\url{https://www.wikidata.org/}}~\cite{wikidataVrandevcic2014}- a general purpose KG and the SEmantic Processing of Security Event Streams CyberSecurity Knowledge Graph (SEPSES CSKG)~\cite{inbook_SEPSESKGs}- a security KG. 
SEPSES CSKG was chosen because it is the most complete security knowledge graph that is publicly available currently and Wikidata was added to assess the ability of STI systems to cope with more general KG.

\subsubsection{Security data sources}
CWE~\cite{vulnerabilities2005common} is a community-developed list of software and hardware weaknesses. It is widely used in cybersecurity, secure coding, and vulnerability assessment. Each CWE entry describes a common type of mistake or flaw that could lead to exploitable security issues. The CWE list is updated three to four times per year to add new and update existing weakness information. CWE is free to use by any organisation, individual for any research, development, and/or commercial purposes, per the CWE Terms of Use. The version used to build the secu-table dataset (CWE v4.8-June 2022) was downloaded from the CWE repository.

CVE~\cite{christey2013common} program aims to identify, define, and catalogue publicly disclosed cybersecurity vulnerabilities. It helps track vulnerabilities that need patching, appear in alerts about active exploitation campaigns, used to automate security into continuous integration/continuous delivery pipelines. Each CVE entry contains structured metadata to help identify, track, and remediate the issue. The version used in this work was downloaded in 2022 from CVE (CVE IDs up to CVE-2022-99999) repository.

\subsubsection{Knowledge Graphs}
A knowledge graph (\(KG=(E, R, T)\)) is a labelled directed multi-graph in which nodes (\(E\)) represent a set of real-world entities, edge types represent relation (\(R\)) between nodes and \(T \subseteq E \times R \times E\) is a set of triples, where each triple (\(e_i, r, e_j\)) represents a directed edge from head entity \(e_i \in E\) to tail entity \(e_j \in E\) via relation \(r \in R\). In this work, we aim to link flat data in security databases to existing KGs so as to add semantics to these tables~\cite{Paulheim2017}. To this end, the world largest general purpose KG Wikidata and a security domain specific KG SEPSES CSKG are used. The Table \ref{tab:KG_comparison} presents a comparison of these KGs.

\begin{table}[!ht]
    \centering
    \small
    \caption{Comparison of KGs used in this work}
    \begin{tabular}{p{2.2cm} p{1cm} p{1.8cm} p{1cm} p{1.4cm} p{1.5cm} p{1.1cm}}
    \toprule
    \textbf{KGs} & \textbf{Year} & \textbf{Domain} & \textbf{Model} & \textbf{Entities} & \textbf{Relations} & \textbf{Types} \\
    \midrule
    Wikidata & 2012 & General knowledge & RDF & 100M & 14B & 300K \\
    SEPSES CSKG  & 2019 & Cyber security & RDF & 3.8M & 479 & --- \\
    \bottomrule
    \end{tabular}
    \label{tab:KG_comparison}
\end{table}

Wikidata~\cite{wikidataVrandevcic2014} is a general purpose, free and open source KG maintained by the Wikimedia Foundation. It aims to store structured data for all sorts of topics, concepts, and objects. Its content is available under a creative commons public domain license. Wikidata helps in knowledge integration by connecting different pieces of information and different datasets, source of standardized identifiers.

The SEPSES CSKG~\cite{inbook_SEPSESKGs} is a cybersecurity KG developed by TU Wien and SBA research group. It integrates and links critical information from publicly available sources. It is constructed using several security data sources such as the Common Weakness Enumeration (CWE) taxonomy, the Common Vulnerabilities and Exposures (CVE) database, the Common Attack Pattern Catalog (CAPEC), and the Security Content Automation Protocol (SCAP). The SEPSES CSKG~\footnote{\url{https://sepses.ifs.tuwien.ac.at/dumps/version/102019/graph000001\_000001.ttl.gz}} was downloaded and used during table annotation.

\subsection{Data curators}
Data curators consisted of Master students and one professor in computer science who are co-authors of this paper. These people have a strong background in semantic web, semantic table interpretation and knowledge graphs. The data curators were divided into two groups: the first group consisted of people responsible for the creation of the tabular dataset and the second group of people responsible for the annotation. The most cumbersome and time consuming task was the data annotation. Thus, each data curator was trained on how to find relevant annotations in the different KGs identified. After the training, a qualification test consists of giving five tables to data curators for curation and evaluating their curation by the expert curators. Expert curators have at least two years experience in semantic table interpretation. They were also responsible for quality review, by checking the data curated.

\subsection{Data collection \& Tables construction}
The current version of the secu-table dataset is built using the CWE and CVE entries. These entries involved a unique number, a short description, what the weakness is and why it matters, common consequences, how the weakness might be exploited, real-world cases and the detection and prevention guidance. The data were downloaded from these data sources into different formats such as JSON, XML, CSV. Thereafter, these data were manually parsed to obtain CSV files.

CVE and CWE provide structured data in formats such as JSON, XML, CSV. Therefore, tables were created by exploiting these structured metadata fields. For instance, CWE metadata (e.g., $"CWE-ID"$, $"Name"$, ... $"Description"$,    $"Related Weaknesses"$, ...) extracted from CWE sources were used to construct tables\footnote{\url{https://huggingface.co/datasets/jiofidelus/SecuTable/blob/main/secutable_v2/ground_truth/tables/table1.csv}} in which $"CWE-ID"$ corresponds to the $ID$ of the underlying software weakness, $"name"$ corresponds to the weakness name, $"Description"$ corresponds to the weakness description, $"Related Weaknesses"$ corresponds to related weakness, etc. The column content was obtained by grouping entries by considering the CWE metadata under each element (e.g., weakness name under the column $"weakness"$).

\subsection{Dataset annotation}
\begin{figure}[!ht]
\caption{Illustration of cell entity annotation, column type annotation, and column property annotation using the SEPSES CSKG}
    \centering
    \includegraphics[scale=0.20]{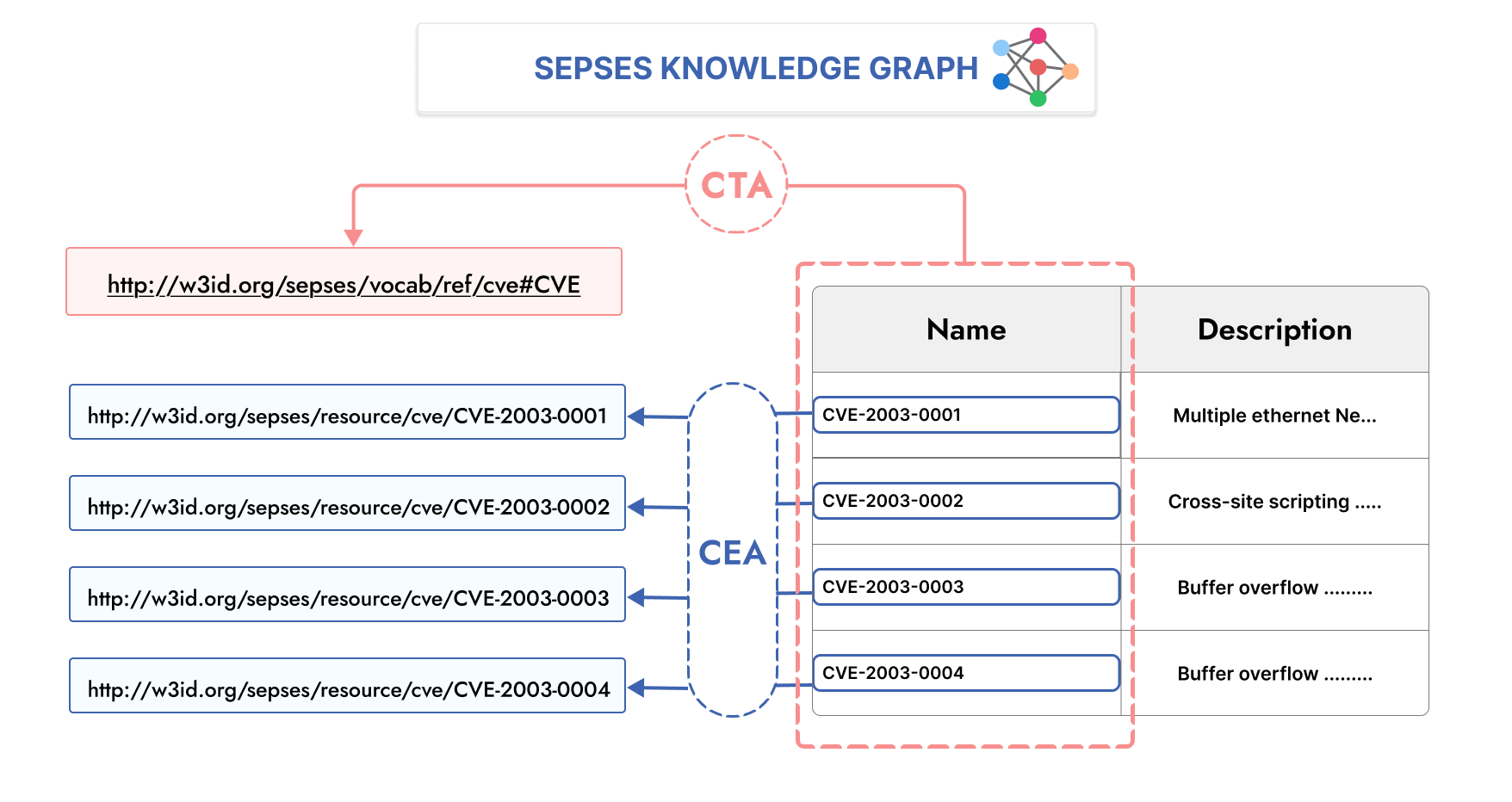}  
    \label{fig:cta_cea_cpa2_illustrations}
\end{figure}

To enhance the understanding of a table, table annotation consists of mapping the table elements to an existing KG. The annotation tasks are illustrated by Fig. \ref{fig:cta_cea_cpa2_illustrations} and defined by equation \ref{eq:anno}.
\begin{itemize}
    \item The Cell Entity Annotation (CEA) task consists of mapping table elements to corresponding entities in the KG. Therefore, given an entity \(e_{tab}\) in the table \(tab\), the goal is to find its corresponding entity \(e_{kg}\) in the KG.
    \item Column Type Annotation (CTA) task consists of mapping columns in tables to its corresponding types in the KG. Thus, given a column \(c_{tab}\) in the table \(tab\), the goal is to find the corresponding type \(t_{kg}\) in the KG.
    \item Column Property Annotation (CPA) task consists of identifying the relation between columns \(c_{tab_i}\) and \(c_{tab_j}\) in table \(tab\) and mapping the latter to corresponding property \(p_{kg}\) in the KG.
\end{itemize}

\begin{equation}
\label{eq:anno}
\begin{split}    
    cea(e_{tab}) = e_{kg}   \\
    cta(c_{tab}) = t_{kg}   \\
    cpa(c_{tab_i}, c_{tab_j}) = p_{kg}    \\
\end{split}
\end{equation}

In the equation \ref{eq:anno}, the \(cea\) function takes as input an entity in the table and find its corresponding annotation in the KG. The \(cta\) function takes as input a table column (composed of entities stored in this column) and finds the types of the elements contained in this column in the KG and the \(cpa\) function takes as input two columns and finds corresponding properties in the KG.

Given that the dataset is being used to evaluate STI during the SemTab challenge in 2025, we need a high quality dataset without errors. Thus, this dataset was manually annotated. It should be noted that manual annotations are generally used to build tabular datasets for evaluating STI~\cite{BiodivTab2021}. In our case, the annotation is performed with two annotators. During the first round, one annotator provide the different links to the KG. To guarantee the quality of the annotation, a second annotator verifies the data annotated.

\begin{lstlisting}[caption={SPARQL query allowing to extract CEA annotations for CWE tables. For each CWE with identifier 5, entities and properties are identified from the knowledge graph and extracted}, label={code:SPARQLCEACWE}, basicstyle=\ttfamily\small, backgroundcolor=\color{gray!10}]
PREFIX rdfs: <http://www.w3.org/2000/01/rdf-schema#>
PREFIX rdf: <http://www.w3.org/1999/02/22-rdf-syntax-ns#>
PREFIX ns2: <http://w3id.org/sepses/resource/cwe/>
SELECT ?cwe ?pre ?obj
WHERE {
?cwe a <http://w3id.org/sepses/vocab/ref/cwe#CWE> .
?cwe ?pre ?obj .
FILTER (STRENDS(STR(?cwe), "CWE-5"))
}
\end{lstlisting}

\begin{lstlisting}[caption={SPARQL query allowing to extract CEA annotations for CVE tables. For each CVE with identifier 5, entities and properties are identified from the knowledge graph and extracted}, label={code:SPARQLCEACVE}, basicstyle=\ttfamily\small, backgroundcolor=\color{gray!10}]
PREFIX ref: <http://w3id.org/sepses/vocab/ref/cve#>
PREFIX rdf: <http://www.w3.org/1999/02/22-rdf-syntax-ns#>

SELECT ?cve ?property ?value
WHERE {
  ?cve ?property ?value .
  FILTER (CONTAINS(STR(?cve), "CVE-2018-6147"))
}
ORDER BY ?property
\end{lstlisting}

\textbf{Annotation using SEPSES CSKG.}
During SEPSES annotation, we deployed the SEPSES CSKG on a local machine using Jena triple store database\footnote{\url{https://jena.apache.org/documentation/fuseki2/}}. Thereafter, we ran SPARQL queries to search for annotations. The code listing \ref{code:SPARQLCEACWE} and \ref{code:SPARQLCEACVE} present the codes that we wrote to retrieve the CEA. After the entities are retrieved, in the case of ambiguity (one query to the KG returns many entities), the contextual information from table rows and columns is leveraged to assess the relevance of candidate annotations and to disambiguate among multiple possible matches.

\begin{figure}
    \centering
    \includegraphics[scale=0.3]{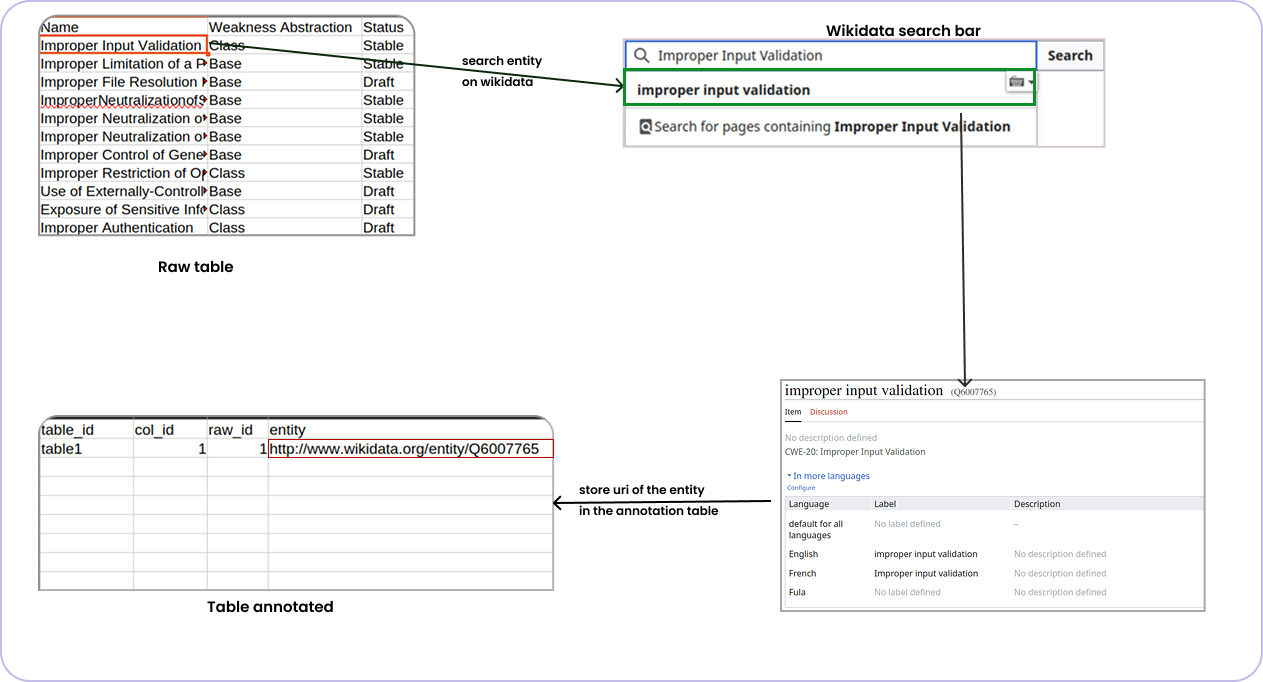}
    \caption{Example of CEA annotation using Wikidata}
    \label{fig:WikidataCEASearch}
\end{figure}

\textbf{Annotation using Wikidata KG.}
Table annotations were identified from Wikidata by manually entering relevant search terms, corresponding to tables elements, entered in the Wikidata search interface (see the illustration of the Fig. \ref{fig:WikidataCEASearch}). Actually, we found that with SPARQL queries, many annotations provided by the query results were not relevant and there were too many ambiguities. Thus, manual search allows us to browse annotations one by one to identify relevant ones. In case of ambiguity (e.g., \textit{"Window Server"} - corresponds to multiple versions and editions in the KG - the search bar of Wikidata provides \textit{"Windows Server 2003 (Q11246)", "Windows Server 2012 (Q11222)", "Windows Server 2008 R2 (Q11226)"}, etc. ), all of which have overlapping labels or aliases. The contextual clues were used from other columns, such as \textit{"Version = 2008 R2"}. Using these clues, the correct entity was selected (\textit{"Windows Server 2008 R2 (Q11226)"}). We also realised that only a small number of tables were providing annotations. The limited number of available annotations from Wikidata resulted in many empty annotation fields, making it difficult to generate reliable CTA and CPA. Consequently, only the CEA was produced in this version of the dataset. 

\subsection{Dataset construction}
The final step of this work consisted of constructing the dataset that will be used for the evaluation of STI. With this dataset, the LLMs should be able to use the table content to detect appropriate annotations. To evaluate the robustness of LLMs for semantic table interpretation, we introduce various types of errors and ambiguities into the dataset using the Pandas library. The dataset was modified as follows:
\begin{itemize}
    \item 20\% data without errors,
    \item 26\% missing context,
    \item 26.6\% misspelling errors,
    \item and 26.26\% annotation errors.
\end{itemize}
The annotation errors consist of labelling data with wrong labels. These controlled errors simulate real-world challenges in security data annotation, helping to benchmark and improve LLMs performance in handling misspelled, ambiguous and incomplete information.

\section{Data record}
\label{sec:datasetDes}
The secu-table dataset is publicly available for research purposes on Hugging Face\footnote{\url{https://huggingface.co/datasets/jiofidelus/SecuTable/}}. Concerning the current release\footnote{\url{https://huggingface.co/datasets/jiofidelus/SecuTable/blob/main/secutable_v2}} (secu-table v2), all data components were downloaded, curated in 2023 for the CEA, CTA and CPA tasks. The dataset contains the following folders:
\begin{itemize}
    \item \textbf{secutable\_v2} is the main folder of the dataset, containing all benchmark components necessary for evaluation and replication. This includes the ground truth annotations, raw tables, and test tables.
    \item \textbf{secutable\_v2/ground\_truth/} contains the ground truth annotations, made available to assist the research community in training, evaluating, and benchmarking STI systems. It includes the following subfolders:
    \begin{itemize}
        \item /sepses/: contains ground truth annotations aligned with the SEPSES CSKG.
        \item /wikidata/: contains ground truth annotations aligned with Wikidata KG.
        \item /table/: contains the raw tables that were annotated using both SEPSES CSKG and Wikidata. These tables can be divided into train and test during STI development.
    \end{itemize}
    \item \textbf{/secutable\_v2/test/tables/} includes the test tables that researchers can use to evaluate and benchmark their systems. The annotations of these tables are hidden so as to be used as the test for STI systems during the SemTab challenges.
\end{itemize}

The dataset content indicates that despite manual annotation, several columns still contain empty entries due to the inherent incompleteness of the underlying KG. KG incompleteness is a well-known limitation affecting virtually all existing KGs~\cite{Paulheim2017}. KG incompleteness is a common challenge, as many entities and relationships are either partially described or entirely missing in real-world KGs. Consequently, the presence of empty cells reflects this real-world incompleteness and provides a realistic scenario in which downstream systems must handle missing or partial annotations. This design ensures that benchmarking and evaluation capture the practical challenges associated with sparse and incomplete KGs.

This work relies on only two security datasources. In future iterations, we plan to extend the dataset to incorporate additional data sources and knowledge graphs (e.g., DBpedia). Given the high cost and scalability issue of manual annotation, we are exploring a semi-automatic approach combining LLMs (e.g., Falcon3-7b-instruct) with human-in-the-loop to verify the quality of the annotations. Therefore, from January onward, new releases of the dataset will occur on a quarterly basis. These releases will be named according to the modification of the dataset: integration of new tables, integration of new KG or integration of new security data sources.

\section{Data Overview}
\label{Data Overview}
The current release of the secu-table dataset consists of 1,554 tables, divided into 76 tables provided as ground truth and 1,478 tables for testing and supports the CEA, CTA and CPA annotations. The ground truth table contains more than 8,900 entities, 55,000 rows, 1000 columns. The test tables contain more than 150k entities, 1M rows and 20k columns. The average number of columns per table is 8.13 and the average number of rows per table is 291.63.

\section{Technical Validation}
\label{sec:datasetEval}
The secu-table\_v2 dataset was constructed for the purpose of the SemTab@ISWC 2025 challenge. This challenge aims to evaluate the capacity of LLMs to successfully annotate security data using Wikidata and SEPSES KGs. This section presents a preliminary evaluation of the dataset using two open source LLMs and one closed source LLM. The code is provided on GitLab\footnote{\url{https://gitlab.com/fidel.jiomekong/secutable}} using the MIT license.

Preliminary evaluation consisted of solving the CEA, CTA and CPA tasks on a subset of the dataset composed of 76 tables (ground truth) using Falcon3-7b-instruct\footnote{\url{https://falconfoundation.ai/}}~\cite{almazrouei2023falcon}, Mistral\footnote{\url{https://mistral.ai/}}~\cite{jiang2023mistral7b} for the open source LLMs and GPT-4o mini for closed source LLM. LLMs used during evaluation were chosen based on their license (open source), competitive performance against other open and closed source LLMs (using Hugging Face LLM Leaderboard) and their integration with the Hugging Face Transformers library. The following hyperparameters were used for Mistral and Falcon: Temperature=0.7, Top-p=0.95, do-sample=true, use-cache=true, max-new-tokens=512, top-k=50. Given that the task is to assess how well models can annotate, the KG was not explicitly provided to the model as input. A two-shot prompting strategy, where the model receives two illustrative examples in the prompt before the question was used. The baseline, the code and the prompts are provided in the dataset documentation so as to allow newcomers to start from somewhere. The LLMs was run on a 12 core server having the following characteristics: RAM=48Go, GPU type: NVIDIA RTX A3090, GPU memory=48Go. 

Semantic table interpretation systems are evaluated based on precision, recall, and F-score~\cite{efthymiou2017matching}. Precision (calculated using \ref{eq:precisionEQ}) is the proportion of correctly predicted annotations out of all predicted annotations. Recall (see equation \ref{eq:recallEQ}) is the proportion of correctly predicted annotations out of all annotations. And the F1-score (see equation \ref{eq:fmesureEQ}).

\begin{equation}
\label{eq:precisionEQ}
    Precision = \frac{\text{relevant annotations}} {\text{total annotations}}
\end{equation}

\begin{equation}
\label{eq:recallEQ}
    Recall = \frac{\text{relevant annotations}} {\text{ground truth annotations}}
\end{equation}

\begin{equation}
\label{eq:fmesureEQ}
    F-score = \frac{2*Precision*Recall}{Precision+Recall}
\end{equation}

\section{Code Availability}
All the source code used during this work is available on GitLab: \url{https://gitlab.com/fidel.jiomekong/secutable} using the MIT license.

\section{Funding}
This research received no specific grant from any funding agency in the public, commercial, or not-for-profit sectors. The work was carried out independently by the authors.

\newpage

\bibliographystyle{unsrt}
\bibliography{0-2025/neurips_2025}

\newpage
\section*{NeurIPS Paper Checklist}


\begin{enumerate}

\item {\bf Claims}
	\item[] Question: Do the main claims made in the abstract and introduction accurately reflect the paper's contributions and scope?
	\item[] Answer: \answerYes{} 
	\item[] Justification: \justificationTODO{The main claims made in the abstract and introduction which is the construction of a security tabular dataset for the evaluation of semantic table interpretation systems accurately reflect the paper's contributions and scope. Actually, the dataset is constructed and evaluated (see sections \ref{sec:methodology}, \ref{sec:datasetDes}, \ref{sec:datasetEval}).}

\item {\bf Limitations}
	\item[] Question: Does the paper discuss the limitations of the work performed by the authors?
	\item[] Answer: \answerYes{}
	\item[] Justification: \justificationTODO{The paper discuss the limitations of the work performed (see section \ref{sec:methodology})}

\item {\bf Theory assumptions and proofs}
	\item[] Question: For each theoretical result, does the paper provide the full set of assumptions and a complete (and correct) proof?
	\item[] Answer: \answerNA{}
	\item[] Justification: \justificationTODO{This is not a theoretical paper}

	\item {\bf Experimental result reproducibility}
	\item[] Question: Does the paper fully disclose all the information needed to reproduce the main experimental results of the paper to the extent that it affects the main claims and/or conclusions of the paper (regardless of whether the code and data are provided or not)?
	\item[] Answer: \answerYes{} 
	\item[] Justification: \justificationTODO{We explained the dataset construction methofology in section \ref{sec:methodology} and evaluations, highlighting hyperparameters in section \ref{sec:datasetEval}}


\item {\bf Open access to data and code}
	\item[] Question: Does the paper provide open access to the data and code, with sufficient instructions to faithfully reproduce the main experimental results, as described in supplemental material?
	\item[] Answer: \answerYes{} 
	\item[] Justification: \justificationTODO{Hugging Face and GitLab links are provided. These resources are published using the MIT license}

\item {\bf Experimental setting/details}
	\item[] Question: Does the paper specify all the training and test details (e.g., data splits, hyperparameters, how they were chosen, type of optimizer, etc.) necessary to understand the results?
	\item[] Answer: \answerYes{} 
	\item[] Justification: \justificationTODO{Details are presented in section \ref{sec:datasetEval}. Additional resources such as annotated data and code description are provided in supplementary materials.}


\item {\bf Experiment statistical significance}
	\item[] Question: Does the paper report error bars suitably and correctly defined or other appropriate information about the statistical significance of the experiments?
	\item[] Answer: \answerNA{} 
	\item[] Justification: \justificationTODO{Experiments results are evaluated with Recall, Precision and F-score metrics}

\item {\bf Experiments compute resources}
	\item[] Question: For each experiment, does the paper provide sufficient information on the computer resources (type of compute workers, memory, time of execution) needed to reproduce the experiments?
	\item[] Answer: \answerYes{} 
	\item[] Justification: \justificationTODO{The details are presented in section \ref{sec:datasetEval}}
    
\item {\bf Code of ethics}
	\item[] Question: Does the research conducted in the paper conform, in every respect, with the NeurIPS Code of Ethics \url{https://neurips.cc/public/EthicsGuidelines}?
	\item[] Answer: \answerYes{} 
	\item[] Justification: \justificationTODO{We have read and understood the NeurIPS code of ethics, and have done our best to conform}

\item {\bf Broader impacts}
	\item[] Question: Does the paper discuss both potential positive societal impacts and negative societal impacts of the work performed?
	\item[] Answer: \answerNA{} 
	\item[] Justification: \justificationTODO{This work consists of evaluating the performance of semantic table annotation systems on domain specific datasets and does not impact the  society at large.}

    
\item {\bf Safeguards}
	\item[] Question: Does the paper describe safeguards that have been put in place for responsible release of data or models that have a high risk for misuse (e.g., pretrained language models, image generators, or scraped datasets)?
	\item[] Answer: \answerYes{} 
	\item[] Justification: \justificationTODO{Since we trained open source LLMs on the dataset built, our work poses no risk of misuse.}
    

\item {\bf Licenses for existing assets}
	\item[] Question: Are the creators or original owners of assets (e.g., code, data, models), used in the paper, properly credited and are the license and terms of use explicitly mentioned and properly respected?
	\item[] Answer: \answerYes{} 
	\item[] Justification: \justificationTODO{We are not shipping the dataset and code from any other existing dataset and code. We did cite open-sourced LLMs and libraries used during this work.}

\item {\bf New assets}
	\item[] Question: Are new assets introduced in the paper well documented and is the documentation provided alongside the assets?
	\item[] Answer: \answerYes{} 
	\item[] Justification: \justificationTODO{The dataset is released on Hugging Face, with included README files. The README file will be updated with more details documentations}


\item {\bf Crowdsourcing and research with human subjects}
	\item[] Question: For crowdsourcing experiments and research with human subjects, does the paper include the full text of instructions given to participants and screenshots, if applicable, as well as details about compensation (if any)?
	\item[] Answer: \answerNA{} 
	\item[] Justification: \justificationTODO{This work does not involve crowdsourcing nor research with human subjects}


\item {\bf Institutional review board (IRB) approvals or equivalent for research with human subjects}
	\item[] Question: Does the paper describe potential risks incurred by study participants, whether such risks were disclosed to the subjects, and whether Institutional Review Board (IRB) approvals (or an equivalent approval/review based on the requirements of your country or institution) were obtained?
	\item[] Answer: \answerNA{} 
	\item[] Justification: \justificationTODO{This work does not involve crowdsourcing nor research with human subjects}

\item {\bf Declaration of LLM usage}
	\item[] Question: Does the paper describe the usage of LLMs if it is an important, original, or non-standard component of the core methods in this research? Note that if the LLM is used only for writing, editing, or formatting purposes and does not impact the core methodology, scientific rigorousness, or originality of the research, declaration is not required.
	\item[] Answer: \answerNA{} 
	\item[] Justification: \justificationTODO{LLMs were used for grammar checking and to polish the quality of the document.}

\end{enumerate}

\end{document}